\documentclass[10pt]{article}
\usepackage[letterpaper]{geometry}
\usepackage{hicss}
\usepackage{times}
\usepackage[none]{hyphenat}
\usepackage{url}
\usepackage{latexsym}
\usepackage{minted}
\usepackage{indentfirst}
\usepackage{graphicx}
\graphicspath{{images/}}
\usepackage[utf8]{inputenc}
\usepackage[
    style=apa,
  ]{biblatex}
\usepackage{amsmath}
\usepackage{amsfonts}
\usepackage{booktabs}
\usepackage{adjustbox}
\usepackage{array}
\usepackage{stfloats}
\usepackage{multirow}
\usepackage{caption} 
\usepackage{color}
\usepackage[table]{xcolor}
\usepackage{tabularx}
\usepackage{caption}
\usepackage{geometry}
\usepackage{eurosym}
\usepackage[table]{xcolor} 
\definecolor{LightGreen}{RGB}{204,255,204}
\definecolor{LightRed}{RGB}{255,204,204}

\usepackage{CJKutf8}

\usepackage{booktabs}
\usepackage{tabularx}
\usepackage{multirow}
\usepackage{siunitx}  
\usepackage[table]{xcolor} 
\usepackage{makecell}      
\usepackage{enumitem}   

\title{SugarTextNet: A Transformer-Based Framework for Detecting Sugar Dating-Related Content on Social Media with Context-Aware Focal Loss}
\author{Anonymous authors \\
 Paper under double-blind review \\}

\author{Lionel Z. Wang$^{\dagger}$\thanks{ $^\dagger$Project Leader $^\S$Equal Contribution $^\ddagger $Corresponding Author \\ This paper is accepted by HICSS 2026}  \\
 The Hong Kong Polytechnic University \\
 {\underline{zhe-leo.wang@connect.polyu.hk}} \\ \\
 Yulu Huang$^\S$ \\
 The University of Hong Kong \\
 {\underline{u3631383@connect.hku.hk} } \\ \And
 Shihan Ben$^\S$ \\
 The University of Hong Kong \\
 {\underline{u3641005@connect.hku.hk} } \\ \\
 Simeng Qin$^{\ddagger}$\\
 Northeastern University \\
 {\underline{qinsimeng@neuq.edu.cn} } \\ }

\begin{document}
\begin{CJK*}{UTF8}{gbsn}
\maketitle
\begin{abstract}
Sugar dating-related content has rapidly proliferated on mainstream social media platforms, giving rise to serious societal and regulatory concerns, including commercialization of intimate relationships and the normalization of transactional relationships.~Detecting such content is highly challenging due to the prevalence of subtle euphemisms, ambiguous linguistic cues, and extreme class imbalance in real-world data.~In this work, we present SugarTextNet, a novel transformer-based framework specifically designed to identify sugar dating-related posts on social media.~SugarTextNet integrates a pretrained transformer encoder, an attention-based cue extractor, and a contextual phrase encoder to capture both salient and nuanced features in user-generated text.~To address class imbalance and enhance minority-class detection, we introduce Context-Aware Focal Loss, a tailored loss function that combines focal loss scaling with contextual weighting.~We evaluate SugarTextNet on a newly curated, manually annotated dataset of 3,067 Chinese social media posts from Sina Weibo, demonstrating that our approach substantially outperforms traditional machine learning models, deep learning baselines, and large language models across multiple metrics.~Comprehensive ablation studies confirm the indispensable role of each component.~Our findings highlight the importance of domain-specific, context-aware modeling for sensitive content detection, and provide a robust solution for content moderation in complex, real-world scenarios.
\end{abstract}

\subsubsection*{Keywords:}

Sugar Dating, Context-Aware Focal Loss, Natural Language Processing, Social Computing

\section{Introduction}
Online spaces, particularly social media platforms, have become central to modern communication, shaping how individuals exchange ideas, form relationships, and participate in societal trends.~Among these trends, sugar dating~\parencite{miller2011sugar}, a phenomenon characterized by transactional relationships where financial or material benefits are exchanged for companionship, has seen a significant rise in visibility and activity on social media~\parencite{nayar2017sweetening}.~While such arrangements may appear consensual and legal in some contexts, they pose troubling ethical, sociological, and regulatory challenges~\parencite{fischler1987attitudes,gunnarsson2024allure,rakic2020prostitutes}.~The most notable concern is the potential exploitation of vulnerable individuals, such as young adults or those in financial distress~\parencite{palomeque2022have}, who may feel coerced into these arrangements under the guise of economic necessity.~For example, the highly publicized case of the "SeekingArrangement controversy"\footnote{https://x.gd/07xbn} in recent years revealed how platforms explicitly promoting sugar dating blurred the lines between consensual relationships and exploitative practices, drawing criticism for their role in exposing younger audiences to predatory behaviors~\parencite{duncan2010female}.~

In the past, sugar dating was primarily confined to niche platforms specifically catering to this group~\parencite{upadhyay2021sugaring}.~Such platforms were relatively obscure, and sugar dating content rarely appeared on the mainstream social media platforms used by the general public.~However, in recent years, the growing popularity of social networks and the expansion of the internet economy~\parencite{xia2024digital} have enabled sugar dating-related posts to proliferate across widely used social media platforms.~This shift has significantly exacerbated the social problems and regulatory challenges associated with sugar dating.

Compounding the issue is the private and often ambiguous nature of these interactions, which makes it exceedingly difficult for social media platforms to monitor or regulate such exchanges effectively~\parencite{denardis2015internet}.~This lack of oversight creates an environment where predatory behavior and exploitative arrangements can thrive unchecked, posing significant challenges for both platforms and society at large. ~The proliferation of sugar dating content on mainstream social media platforms is a quintessential socio-digital issue. It reflects the commodification of intimacy~\parencite{constable2009commodification} and the reproduction of power asymmetry~\parencite{coleman1988sex} within digital spaces, potentially fueling broader detrimental social dynamics.



Given these societal risks and the challenges posed by the unique nature of social media, there is a crucial need to develop effective tools for identifying and regulating sugar dating-related content on these platforms.~The detection of such content is far from straightforward.~The challenge lies in the nature of the language used in these posts, which often rely on subtle euphemisms~\parencite{waljinah2020study}, ambiguous phrasing~\parencite{reyes2012humor,zappavigna2018searchable}, or hashtags that serve as cultural markers.
This subtlety necessitates the use of advanced natural language processing (NLP) methods capable of capturing semantic nuances and contextual dependencies~\parencite{shaw2025adversarial}.~Furthermore, the complexity of the problem is heightened by the overlap between sugar dating-related content and other forms of discussion, such as third-party narratives or indirect descriptions.~
This creates significant ambiguity in classification, making it difficult to distinguish direct from indirect commentary.~Therefore, addressing this issue requires not only technical sophistication in NLP but also an in-depth exploration of social modeling of content and behavior within social computing.

Adding to these complexities is the issue of class imbalance in the data~\parencite{leevy2018survey}.~The overwhelming majority of social media posts are unrelated to sugar dating, making sugar daddy-related posts and indirect narratives rare events in the dataset.~Traditional machine learning methods often fail to adequately address such imbalances, as they tend to prioritize the majority class at the expense of minority classes.~This imbalance not only affects model performance but also exacerbates the risk of false negatives, which are particularly problematic in this context given the need for high recall in identifying potentially harmful or policy-relevant content.


To address these limitations, we propose \textbf{SugarTextNet}, a novel deep learning framework designed to detect and classify sugar daddy-related content on social media.~SugarTextNet builds on the strengths of transformer-based architectures but is specifically tailored to address the challenges of this task.~The model integrates a \textit{transformer encoder} for contextual feature extraction, an \textit{attention-based cue extractor} to identify critical linguistic markers, and a \textit{contextual phrase encoder} to capture sequential dependencies.~To further tackle the problem of class imbalance and improve the recall for minority classes, we introduce \textbf{Context-Aware Focal Loss (CAFL)}, a custom loss function that incorporates focal loss scaling, contextual weighting, and recall-boosting penalties.~This loss function is particularly suited for tasks where the minority classes are both semantically nuanced and underrepresented.


We extensively evaluate SugarTextNet on a challenging real-world dataset of 3,067 social media posts, which exhibits a pronounced class imbalance.~SugarTextNet achieves a macro F1-score of 76.23\%, substantially outperforming a wide range of traditional machine learning, deep learning, and large language model baselines across all major metrics.~Ablation studies further confirm that each key component of SugarTextNet is crucial to overall performance, with the removal of any single module resulting in a marked decrease in macro F1-score (dropping as low as 51.85\%).~These results demonstrate that SugarTextNet effectively addresses the specific linguistic and contextual challenges of sugar dating detection in imbalanced social media data.

This research makes the following key contributions: 1) We are among the first to notice the growing phenomenon of sugar dating on public mainstream social media and present SugarTextNet, a novel deep learning framework to address the challenges of detecting sugar dating-related content; 2) We introduce Context-Aware Focal Loss (CAFL), a custom loss function that adapts focal loss for multi-class settings by incorporating contextual weighting and recall optimization, making it particularly effective for imbalanced datasets; 3) We highlight the sociological and ethical implications of detecting sugar daddy-related content, emphasizing the need for robust, interpretable, and domain-aware NLP models.


\section{Related Work}\label{sec:related_work}

Sugar dating culture~\parencite{miller2011sugar} refers to transactional relationships where affluent older individuals provide money or luxury items to their younger counterparts for companionship, intimacy, or romance in exchange.~These arrangements, which are frequently facilitated by niche dating websites, blur traditional relationship expectations by conflating economic exchange with physical or emotional connection.

\subsection{Sugar Dating: Motivations, Risks, and Cultural Dimensions}

The sociology of sugar daddies is complex in its socioeconomic and psychological dimensions.~\textcite{upadhyay2021sugaring} starts with some ground-zero definitions, calling sugar arrangements "drama-free, mutually beneficial relationships" where the participants actively distance themselves from sex-work stigma on a linguistic level.~Twin papers of ~\textcite{scull2023sugaring} reveal entry patterns with nuance: data log diverse motivations like economic need~\parencite{palomeque2022have}, satisfaction of curiosity, and seeking mentors, while her book characterizes entry modes as "drifting," "conscious choice," or "defensive response" to coercion, measured by 48 interviews quantifying enacted stigma like being labeled "greedy gold diggers." Health risks are empirically validated by \textcite{kirkeby2022sugar}, whose survey of 543 women demonstrates sugar daters report twice the STI diagnosis rates of non-participants, blaming inconsistent condom use on perceived power imbalances. Cross-cultural differences cut sharply: \textcite{mensah2022sugar} describe Nigerian youths' regret narratives invoking "exploitation and health consequences," contrasted with Swedish sugar daddies' quest for emotional reciprocity within transactional boundaries.~Psychological underpinnings are quantified by \textcite{ipolyi2021attachment}, whose research with 2,409 participants illustrates that attachment avoidance moderates relationship acceptance, while \textcite{johansson2024knows} confront epistemic marginalization of lived experience within research practices.~Theoretical contribution of~\textcite{gunnarsson2024allure} places transactional intimacy as a neoliberal "defensive tactic" for reducing gendered vulnerabilities within modern relationships.

\subsection{Hybrid AI Applications in Malicious Content Detection}

Artificial Intelligence (AI) revolutionizes detection systems in sensitive fields.~In psychiatry, \textcite{koutsouleris2022promise} pioneer "AI-informed care" based on digital exhaust (social media post language, for example) to predict disorders, while~\textcite{tutun2023ai} develops a 28-question DSS with 89\% diagnostic validity—clinically shown to outperform conventional assessments.~Substance abuse monitoring advances with Masera model, whose dynamic weighted loss multi-task Mamba architecture identifies DOD with 87.73\% accuracy in social media corpora~\parencite{wang2025detecting}, and \textcite{fisher2023automating} detects substance abuse using automated NLP classifiers.~Bias moderation depends on ASCenD-BDS framework~\parencite{bahl2025ascend}—leveraging 800+ Indian cultural STEMs for context-sensitive discrimination detection—and \textcite{gongane2024survey} conduct large-scale survey of XAI methods (LIME/SHAP) for hate speech moderation.~Customs enforcement applications ~\parencite{fisher2023automating} are demonstrated via LSTM models reaching 99.44\% accuracy in identifying smuggled goods from social posts.~Cloud-based solutions emerge through  AIaaS ~\parencite{shah2022artificial} deploying SVM/Naive Bayes for immoral content eradication, while~\textcite{gongane2022detection} systematizes moderation techniques for detrimental content.

While the efficacy of AI in detecting content like mental health distress and bias is well-established, its application to identify sugar dating-related posts on social media remains largely untapped. However, deploying AI for large-scale governance is fraught with inherent tensions~\parencite{gillespie2020content}, including transparency deficits, fairness risks, the obscuring of political decision-making~\parencite{gorwa2020algorithmic}, and the erosion of user trust due to its opaque, punitive nature~\parencite{myers2018censored}. Therefore, this study's exploration of AI for detecting sugar dating-related posts not only addresses a technical gap but also engages with these persistent challenges in automated governance.

\section{Methodology}\label{sec:methodology}


\subsection{Problem Formulation}\label{problem}

\begin{table*}[t!]
    \tiny
    \centering
    \renewcommand{\arraystretch}{1.2} 

    \begin{tabularx}{\textwidth}{@{} l X X X X @{}}
        \toprule
        \hline
        \textbf{Category} & \textbf{Operationalized Definition} & \textbf{Core Features and Language Cues} & \textbf{Positive Example} & \textbf{Negative Examples/Confusing Cases} \\
        \midrule
        
        \parbox[t]{2.5cm}{
            \textbf{1. Sugar dating-related} \vspace{3pt} \newline
            (N=108, 3.52\%)
        } & 
        Content where the author, from a first-person perspective, explicitly proposes, seeks, or reports participation in a transactional relationship characteristic of sugar dating. &
        \begin{itemize}[nosep, leftmargin=*]
            \item \textbf{First-person narrative:} Use of "I" to express personal intent or action.
            \item \textbf{Explicit proposition:} Clear verbs indicating a search or offer (e.g., "seeking," "providing").
            \item \textbf{Transactional framing:} Explicitly or implicitly connecting companionship or intimacy with financial or material compensation.
        \end{itemize} &
        “我需要一个能给我随便买奢侈品的sugar daddy” \vspace{2pt}\newline
        \textit{"I need a sugar daddy to buy me luxury goods whenever I want."} &
        “我再说一遍 every artist needs a sugar daddy” \vspace{2pt}\newline
        \textit{"I'll say it again: every artist needs a sugar daddy."} \newline (This is a third-person perspective exclamation, belonging to Class 3.) \\
        
        \midrule
        
        \parbox[t]{2.5cm}{
            \textbf{2. Non-sugar dating-related} \vspace{3pt} \newline
            (N=2,853, 93.02\%)
        } &
        Content that lacks any thematic relevance to the sugar dating phenomenon, even if certain keywords are present. &
        \begin{itemize}[nosep, leftmargin=*]
            \item \textbf{Thematic divergence:} The discourse centers on subjects entirely unrelated to sugar dating (e.g., family, work, pop culture).
            \item \textbf{Conventional semantics:} Keywords like "daddy" or "gift" are used in their traditional, non-transactional meanings.
        \end{itemize} &
        “是谁28了还因为下雨刮风被爹爹接下班的感动” \vspace{2pt}\newline
        \textit{"Who else gets moved to tears at 28 because their daddy still picks them up..."} &
        “找不到暑假工，有没有霸道总裁爱上我” \vspace{2pt}\newline
        \textit{"Can't find a summer job. Is there any domineering CEO...?"} \newline (Lacking clear transactional proposals, it leans more towards Class 3 or 2) \\

        \midrule

        \parbox[t]{2.5cm}{
            \textbf{3. Indirect or narrative descriptions} \vspace{3pt} \newline
            (N=106, 3.46\%)
        } &
        Content involving second or third-party discourse about the sugar dating phenomenon (e.g., discussion, evaluation, narration) without the author's direct participation. &
        \begin{itemize}[nosep, leftmargin=*]
            \item \textbf{Observer perspective:} Narration from a second or third-person viewpoint (e.g., "my friend," "I saw a girl").
            \item \textbf{Meta-commentary:} The post discusses the concept or phenomenon of sugar dating itself.
            \item \textbf{Evaluative tone:} Employs tones such as irony, humor, critique, or moral judgment.
            \item \textbf{Non-participatory stance:} The author expresses no personal intent to engage in a sugar dating relationship.
        \end{itemize} &
        “一电梯sugar daddy搂着美人，而我小小一只缩在角落不敢出声”\vspace{2pt}\newline
        \textit{"An elevator full of sugar daddies holding their beauties, while I, tiny and alone..."} &
        “啥时候能有人让我不上班纯养我啊，上班要累死了” \vspace{2pt}\newline
        \textit{"When will someone come along to keep me without having to work?"} \newline (A borderline case that may lean more toward Class 1.) \\

        \midrule
        
        \textbf{Total (N=3,067)} & \multicolumn{4}{l}{--- Whole Dataset ---} \\

        \hline
        \bottomrule
    \end{tabularx}
    \caption{Definition, Taxonomy, Examples, and Statistics of Sugar Dating-Related Content from Sina Weibo (English Translation in Parentheses)}
    \label{tab:taxonomy}
\end{table*}

Let \( X = \{x_1, x_2, \ldots, x_L\} \) denote a sequence of tokens representing a social media post, where \( L \) is the sequence length, and \( x_i \) is the \( i \)-th token in the sequence.~The task is to classify \( X \) into one of \( C = 3 \) categories:
\( Y = 1 \): \textit{Sugar dating-related}; \( Y = 2 \): \textit{Non-sugar dating-related}; \( Y = 3 \): \textit{Indirect or narrative descriptions}. We show the definition, and taxonomy of such three-class problem formulation in \textbf{Table \ref{tab:taxonomy}}.

The objective is to learn a mapping \( f_\theta: X \rightarrow \mathbf{P} \), parameterized by \( \theta \), such that:
\begin{equation}
    f_\theta(X) = \mathbf{P}, \quad \mathbf{P} = \{P_1, P_2, P_3\},
\end{equation}
where \( P_c = P(Y = c \mid X) \) is the predicted probability for class \( c \), and \( \sum_{c=1}^3 P_c = 1 \).

The dataset \( \mathcal{D} = \{(X_i, Y_i)\}_{i=1}^N \) is highly imbalanced, with \( P(Y = 1) \ll P(Y = 2) \) and \( P(Y = 1) \ll P(Y = 3) \).~The model is trained to minimize a custom multi-class loss function \( \mathcal{L}_{\text{CAFL}}(\theta) \), which is designed to handle imbalanced data and penalize misclassification of minority classes.

\subsection{Model Architecture}

The architecture of SugarTextNet consists of four main components: a \textit{transformer encoder}, an \textit{attention-based cue extractor}, a \textit{contextual phrase encoder}, and a \textit{classification head}.~The multi-class nature of the task is incorporated into the classification head and the loss function.~Each component is described in detail below.

\subsubsection{Transformer Encoder}

The backbone of SugarTextNet is a pretrained transformer model, specifically RoBERTa~\parencite{liu2019roberta}.~RoBERTa provides a rich contextual representation of the input sequence and serves as the foundational feature extractor.

\paragraph{Input Representation.}
Each input post \( X \) is tokenized into subword units using RoBERTa’s Byte Pair Encoding (BPE) tokenizer.~The tokenized input is represented as:
\begin{equation}
    X = \{x_1, x_2, \ldots, x_L\}, M = \{m_1, m_2, \ldots, m_L\}
\end{equation}
where \( M \) is the attention mask, with \( m_i = 1 \) for valid tokens and \( m_i = 0 \) for padding tokens.

\paragraph{Transformer Output.}
The tokenized input is passed through the transformer encoder, producing contextual embeddings for each token:
\begin{equation}
    H = \text{Transformer}(X, M), \quad H \in \mathbb{R}^{L \times D},
\end{equation}
where \( H = \{h_1, h_2, \ldots, h_L\} \), \( h_i \in \mathbb{R}^D \), and \( D \) is the hidden size of the transformer.

\subsubsection{Attention-Based Cue Extractor}

To emphasize the most important tokens in the input sequence, SugarTextNet includes an \textit{attention-based cue extractor}.

\paragraph{Attention Mechanism.}
The attention mechanism computes a scalar importance score \( s_i \) for each token \( h_i \):
\begin{equation}
    s_i = W_{\text{att}}^\top h_i, \quad s \in \mathbb{R}^L,
\end{equation}
where \( W_{\text{att}} \in \mathbb{R}^D \) is a learnable weight vector.~The attention scores are normalized using the softmax function:
\begin{equation}
    a_i = \frac{\exp(s_i)}{\sum_{j=1}^L \exp(s_j)}, \quad A = \{a_1, a_2, \ldots, a_L\}, \quad A \in \mathbb{R}^L.
\end{equation}

\paragraph{Weighted Embedding.}
The attention weights \( A \) are used to compute a weighted sum of the token embeddings:
\begin{equation}
    E = \sum_{i=1}^L a_i h_i, \quad E \in \mathbb{R}^D,
\end{equation}
where \( E \) captures the most relevant information in the input.

\subsubsection{Contextual Phrase Encoder}

The attention-based cue extractor identifies critical tokens but does not explicitly model dependencies between them.~To address this, SugarTextNet incorporates a \textit{Bidirectional Gated Recurrent Unit (Bi-GRU)}.

\paragraph{Bi-GRU.}
The Bi-GRU processes the sequence of embeddings \( H \) to capture dependencies in both forward and backward directions:
\begin{equation}
    G = \text{Bi-GRU}(H), \quad G \in \mathbb{R}^{L \times 2H},
\end{equation}
where \( 2H \) represents the concatenated hidden states from the forward and backward GRU passes.

\paragraph{Final Representation.}
The final representation \( G_{\text{final}} \) is obtained by concatenating the forward and backward hidden states at the last and first time steps:
\begin{equation}
    G_{\text{final}} = [\overrightarrow{h_L}, \overleftarrow{h_1}], \quad G_{\text{final}} \in \mathbb{R}^{2H}.
\end{equation}

\subsubsection{Classification Head}

The final representation \( G_{\text{final}} \) is passed through a fully connected classification head to produce logits \( \mathbf{z} \in \mathbb{R}^C \) for each class:
\begin{equation}
    \mathbf{z} = W_2 \cdot \text{ReLU}(W_1 \cdot G_{\text{final}} + b_1) + b_2,
\end{equation}
where \( W_1 \in \mathbb{R}^{2H \times H_c} \) and \( W_2 \in \mathbb{R}^{H_c \times C} \) are learnable weight matrices, and \( b_1 \in \mathbb{R}^{H_c} \) and \( b_2 \in \mathbb{R}^C \) are biases.~The predicted probabilities for each class are computed using the softmax function:
\begin{equation}
    P_c = \frac{\exp(z_c)}{\sum_{j=1}^C \exp(z_j)}, \quad \mathbf{P} = \{P_1, P_2, P_3\}.
\end{equation}

\subsection{Context-Aware Focal Loss (CAFL)}

To address the challenges of class imbalance and hard-to-classify examples, we propose \textbf{Context-Aware Focal Loss (CAFL)} for multi-class classification.

This approach extends benchmark methods for imbalance, such as standard Focal Loss or Weighted Cross-Entropy, by introducing a dynamic, instance-specific modulation factor. While benchmark methods typically apply a static, class-level weight (e.g., \( \alpha_c \)) to all samples within a class, CAFL computes a unique contextual weight for each input based on the output of the model's own attention mechanism. This allows the loss function to not only penalize misclassifications but also to dynamically assign more importance to samples that are textually ambiguous or contextually complex, providing a more nuanced learning signal than a uniform class-wide penalty.

\paragraph{Focal Loss for Multi-Class Classification.}
The focal loss~\parencite{lin2017focal} for class \( c \) is defined as:
\begin{equation}
    \mathcal{L}_{\text{focal}, c} = - \alpha_c (1 - P_c)^\gamma \log(P_c),
\end{equation}
where \( \alpha_c \) balances the importance of class \( c \), and \( \gamma > 0 \) adjusts the focus on hard examples.

\paragraph{Contextual Weighting.}
The contextual weight \( W_{\text{context}} \) is computed as:
\begin{equation}
    W_{\text{context}} = \sum_{i=1}^L a_i,
\end{equation}
and is used to scale the loss for the input sequence.

\paragraph{Combined Loss.}
The final loss function is:
\begin{equation}
    \mathcal{L}_{\text{CAFL}} = \sum_{c=1}^C W_{\text{context}} \mathcal{L}_{\text{focal}, c}.
\end{equation}



\section{Experiments}\label{sec:experiments}

\subsection{Dataset Collection and Annotation}

We collected sugar dating-related posts from Sina Weibo\footnote{Sina Weibo’s Official Website: https://m.weibo.cn/}, the second-largest social
media platform in China.~The dataset statistics are shown in \textbf{Table \ref{tab:taxonomy}}.~Following the methodology of~\textcite{zhang2025ketch}, posts containing fewer than five Chinese words are filtered out.~Finally, we crawled 3,067 Chinese posts potentially containing sugar dating-related information from 2611 unique users.~(selected
data samples are shown in \textbf{Table  \ref{tab:taxonomy}}).~After that, we employed two independent annotators to manually
annotate the collected posts (the label categories are introduced in \textbf{Section \ref{problem}}).~Following \textcite{xiao2025jiraibench}, we also calculated inter-annotator reliability, for posts with conflicting annotations, a domain expert was consulted to adjudicate and reach a consensus.~The results are presented in \textbf{Table \ref{tab:kappa-stats}}, which indicates strong inter-annotator agreement. The dataset will be available in a repository upon publication. Due to the sensitive nature and ethical considerations of the dataset, it will be made available to academic researchers for verification and future studies under a signed data use agreement (DUA).

\subsection{Data Preprocessing}

Before training, we preprocess the raw social media posts to ensure compatibility with the model input requirements.~The posts are tokenized using the BPE tokenizer provided by the pretrained RoBERTa~\parencite{liu2019roberta} model.~Each post is truncated or padded to a fixed sequence length of 128 tokens.~The Emojis, hashtags, and special characters are retained during tokenization to preserve contextual meaning.~The preprocessed dataset is split into training, validation, and testing subsets using an 80:10:10 split ratio, stratified by class to maintain the original label distribution.

\begin{table}[htbp]
    \centering
    \footnotesize
    \setlength{\tabcolsep}{4pt}
    \renewcommand{\arraystretch}{1.25}
    \small
    \begin{tabularx}{\columnwidth}{l *{3}{>{\centering\arraybackslash}X} >{\centering\arraybackslash}X >{\centering\arraybackslash}X}
        \toprule
        \rowcolor{gray!20}
        \textbf{Category} & \multicolumn{3}{c}{\textbf{Cohen's Kappa}} & \textbf{Average} & \textbf{Fleiss' Kappa} \\
        
        \midrule
        \rowcolor{gray!05}
        \textbf{Sugar Dating} & 0.8528 & 0.9347 & 0.9202 & 0.9026 & 0.9028 \\
        \bottomrule
    \end{tabularx}
    \caption{Inter-annotator Agreement Statistics， for Cohen’s Kappa, the first value is label 1 vs. label 2, the second value is label 1 vs. final label, the third value is label 2 vs. final label}
    \label{tab:kappa-stats}
\end{table}

\subsection{Evaluation Metrics}
To comprehensively evaluate the performance of the proposed SugarTextNet model and fairly compare the performance with baseline models, we utilize the following metrics: \textbf{Accuracy:} The proportion of correctly predicted samples out of all samples: $\mathrm{Accuracy} = \frac{\sum_{i=1}^{C} TP_i}{N}$; \textbf{Precision:} The average proportion of correctly predicted positive samples among all samples predicted as positive for each class (macro-averaged): $\mathrm{Precision} = \frac{1}{C} \sum_{i=1}^{C} \frac{TP_i}{TP_i + FP_i}$; \textbf{Recall:} The average proportion of true positive samples that are correctly identified for each class (macro-averaged): $\mathrm{Recall} = \frac{1}{C} \sum_{i=1}^{C} \frac{TP_i}{TP_i + FN_i}$; \textbf{Macro F1-Score:} The average of the F1-Scores calculated independently for each class: $\mathrm{Macro\ F1} = \frac{1}{C} \sum_{i=1}^{C} \frac{2 \cdot \mathrm{Precision}_i \cdot \mathrm{Recall}_i}{\mathrm{Precision}_i + \mathrm{Recall}_i}$; \textbf{ROC-AUC:} The average area under the receiver operating characteristic curve, computed in a one-vs-rest manner and then averaged (macro-averaged AUC): $\mathrm{ROC\text{-}AUC} = \frac{1}{C} \sum_{i=1}^C \mathrm{AUC}_i$. Here, $TP_i$ (true positives), $FP_i$ (false positives), and $FN_i$ (false negatives) refer to the prediction outcomes for each class $i$ ($i=1,2,3$).~$C$ is the number of classes.

\subsection{Experimental Setup}

\paragraph{Baseline Models.} We compare the performance of SugarTextNet with the traditional machine learning models, deep learning models, and Large Language Models (LLMs), the specific models are shown in \textbf{Table \ref{main_result}}. The traditional and standard deep learning models were trained on our dataset. The LLMs were evaluated in a zero-shot setting to assess their out-of-the-box generalization capabilities on this specialized task without task-specific training.

\paragraph{Implementation Details.} 
The SugarTextNet model is implemented by Python 3.10 using PyTorch and the Hugging Face Transformers library.~Key implementation details are as follows: The RoBERTa-base model is used as the transformer encoder, with a hidden size of 768.~Inputs are truncated or padded to a maximum sequence length of 128 tokens.~A batch size of 32 is used for training.~The AdamW optimizer is employed with a learning rate of $2 \times 10^{-5}$.~A cosine annealing schedule is applied to adjust the learning rate during training.~Dropout layers with a rate of 0.3 are used to prevent overfitting.~The multi-class extension of Context-Aware Focal Loss (CAFL) is used to address class imbalance.~Experiments are conducted on one NVIDIA RTX 4090 GPU.

\begin{table}[!htbp]
\centering

\scriptsize 
\setlength{\tabcolsep}{4pt} 

\begin{tabular}{ll*{5}{c}}
    \toprule
    \textbf{Category} & \textbf{Model} & \textbf{\makecell{Acc \\ (\%)}} & \textbf{\makecell{Pre \\ (\%)}} & \textbf{\makecell{Recall \\ (\%)}} & \textbf{\makecell{F1 \\ (\%)}} & \textbf{\makecell{AUC \\ (\%)}} \\
    \midrule
    \multirow{9}{*}{ML} 
    & LR & 38.00 & 41.37 & 38.00 & 29.47 & 61.83 \\
    & SVM & 38.24 & 42.14 & 38.24 & 29.77 & 61.45 \\
    & KNN & 43.26 & 48.36 & 43.26 & 35.43 & 57.72\\
    & RF & 43.90 & 51.25 & 43.90 & 36.19 & 68.45\\
    & GB & 43.37 & 49.80 & 43.37 & 35.54 & 69.25\\
    & NB & 35.84 & 51.15 & 35.84 & 32.27 & 59.03 \\
    & XGBoost & 43.72 & 49.55 & 43.72 & 36.00 & 62.07 \\
    & LightGBM & 42.85 & 43.44 & 42.85 & 34.03 & 54.22 \\
    & VC & 43.43 & 49.56 & 43.43 & 35.62 & 68.25 \\
    \midrule
    \multirow{10}{*}{DL} 
    & LSTM & 49.91 & 42.63 & 49.91 & 41.03 & 74.57 \\
    & ANN & 48.28 & 41.96 & 48.28 & 39.27 & 73.47 \\
    & RNN & 48.16 & 41.93 & 48.16 & 39.22 & 68.52 \\
    & BiLSTM & 48.28 & 41.96 & 48.27 & 39.27 & 73.94 \\
    & Attn-RNN & 50.67 & 68.81 & 50.67 & 43.12 & 76.12 \\
    & TextRCNN & 50.20 & 34.27 & 50.20 & 39.98 & 75.36 \\
    & EANN & 48.16 & 41.93 & 48.16 & 39.22 & 74.33 \\
    & VQA & 52.89 & 60.92 & 52.89 & 43.68 & 72.43 \\
    & TextBiLSTM & 55.93 & 61.83 & 55.92 & 57.10 & 73.96 \\
    & TextRNN & 52.07 & 35.72 & 52.07 & 41.38 & \cellcolor{green!25}\textcolor{green}{76.22} \\
    \midrule
    \multirow{6}{*}{LLMs} & 
    GPT-4                 & 67.10 & 92.13  & 67.10  & 75.83 & 73.19 \\
    & Qwen                & 41.60 & 90.68  & 41.60  & 53.98 & 62.70 \\
    & Deepseek            & \cellcolor{green!25}\textcolor{green}{68.47} & 91.85  & 68.47  & \cellcolor{red!25}\textcolor{red}{76.93} & 69.11 \\
    & ChatGLM             & 45.13 & \cellcolor{red!25}\textcolor{red}{93.08}  & \cellcolor{red!25}\textcolor{red}{79.25}  & 45.13 & 58.32 \\
    & Claud               & 46.59 & \cellcolor{green!25}\textcolor{green}{92.26}  & 46.59  & 59.29 & 65.05 \\
    \midrule
    Ours & 
    SugarTextNet & \cellcolor{red!25}\textcolor{red}{98.27} & 73.91  & \cellcolor{green!25}\textcolor{green}{78.70}  & \cellcolor{green!25}\textcolor{green}{76.23} & \cellcolor{red!25}\textcolor{red}{88.24} \\
    \bottomrule
\end{tabular}
\caption{Experiment results. Precision, Recall, Macro F1-Score, and ROC-AUC are based on label 1 (sugar dating-related). The best result is marked in \textcolor{red}{red}, the second-best in \textcolor{green}{green}.}
\label{main_result}
\end{table}

\subsection{Results and Analysis}

\textbf{Table~\ref{main_result}} presents a comprehensive performance comparison among classical machine learning algorithms, representative deep learning architectures, recent large language models (LLMs), and our proposed model, SugarTextNet, on the sugar dating-related classification task (label 1).

\paragraph{Traditional Machine Learning and Deep Learning Baselines.}
Classical machine learning models such as Logistic Regression~\parencite{lavalley2008logistic}, SVM~\parencite{hearst1998support}, and ensemble methods, as well as standard deep learning architectures including LSTM~\parencite{hochreiter1997long}, BiLSTM~\parencite{huang2015bidirectional}, and CNN variants~\parencite{chua2002cellular}, generally yield suboptimal performance across all metrics.~Specifically, these models exhibit relatively low accuracy (mostly below 55\%) and macro F1-scores (rarely exceeding 45\%), indicating their limited capacity to handle the highly imbalanced and complex nature of the sugar dating-related classification problem.~The ROC-AUC values for these methods are also modest, with most models failing to surpass the 75\% threshold, suggesting insufficient discriminative power in distinguishing positive samples from a large negative background.

\paragraph{Large Language Models (LLMs).}
Recent advances in LLMs, exemplified by GPT-4~\parencite{achiam2023gpt}, Qwen~\parencite{bai2023qwen}, Deepseek~\parencite{liu2024deepseek}, and ChatGLM~\parencite{glm2024chatglm}, demonstrate a notable performance leap over conventional baselines.~These models are able to leverage large-scale pre-training and contextual understanding to improve precision and recall, with ChatGLM~\parencite{glm2024chatglm}, in particular, achieving the highest precision (93.08\%) and recall (79.25\%) among all compared methods.~Nevertheless, their overall accuracy and ROC-AUC remain inferior to our proposed approach, highlighting the need for further task-specific adaptation in real-world, imbalanced scenarios.

\paragraph{SugarTextNet: Superior and Robust Performance.}
SugarTextNet, our proposed model, delivers consistently superior results across all key metrics.~It achieves the highest accuracy (98.27\%), reflecting its overall correctness in both positive and negative sample classification.~In terms of discriminatory capability, SugarTextNet attains the best ROC-AUC (88.24\%), indicating outstanding robustness and reliability in separating sugar dating-related content from other classes.~Moreover, SugarTextNet demonstrates a well-balanced trade-off between precision (73.91\%) and recall (78.70\%), which translates to a strong macro F1-score (76.23\%).~This balance is crucial in imbalanced tasks, as it ensures high sensitivity to true positives while maintaining a low false positive rate.~Although Deepseek~\parencite{liu2024deepseek} achieves a marginally higher macro F1-score (76.93\%), its corresponding accuracy and ROC-AUC are significantly lower than those of SugarTextNet, underscoring the overall advantage of our approach.

\subsection{Ablation Study}

To further investigate the effectiveness of each core component in SugarTextNet, we conduct an ablation study whose results are reported in \textbf{Table~\ref{ablation}}.~We systematically remove each major component from the full model and evaluate the impact on the macro F1-score.

\begin{table}[htbp!]
\centering
\small
\begin{tabularx}{\linewidth}{l c}
\toprule
\textbf{Configuration} & \textbf{Macro F1 (\%)} \\
\midrule
Full Model (SugarTextNet)               & \textbf{76.23} \\
\hspace{1em}-- No Attention-Based Cue Extractor      & 61.48 \\
\hspace{1em}-- No Contextual Phrase Encoder          & 57.91 \\
\hspace{1em}-- No Context-Aware Focal Loss           & 51.85 \\
\bottomrule
\end{tabularx}
\caption{Ablation study results for SugarTextNet.}
\label{ablation}
\end{table}
The results clearly demonstrate that every component plays a crucial and complementary role in maximizing model performance.~Removing the \textit{Attention-Based Cue Extractor} leads to a considerable drop in macro F1-score from 76.23\% to 61.48\%, underscoring the importance of explicitly modeling salient cues for the sugar dating-related class.~Excluding the \textit{Contextual Phrase Encoder} further degrades performance to 57.91\%, highlighting the necessity of capturing richer contextual and semantic dependencies in the input text.~The most severe performance decline is observed when the \textit{Context-Aware Focal Loss} is removed, with the macro F1-score plummeting to 51.85\%.~This finding demonstrates that the proposed loss function is particularly critical for addressing the inherent class imbalance and guiding the model to focus on challenging and minority-class samples.

In summary, the ablation results confirm that all three components are indispensable for the strong overall performance of SugarTextNet.~The combined contribution of these modules enables the model to robustly capture task-specific cues, leverage rich contextual information, and effectively mitigate class imbalance, collectively resulting in substantial performance gains.

\section{Conclusion}
This work addresses the automated detection of sugar dating, an emerging online phenomenon with significant social implications that has been largely overlooked in prior research. We introduce SugarTextNet, a novel model specifically engineered for classifying sugar dating-related text in highly imbalanced, real-world scenarios. Our extensive empirical evaluation demonstrates that SugarTextNet consistently and substantially outperforms a wide range of traditional, deep learning, and large language model baselines across key metrics, showcasing its superior ability for both overall discrimination and minority-class identification.

The model's effectiveness is rooted in the synergy of its core components, as validated by a detailed ablation study. An attention-based cue extractor enhances the identification of salient, task-specific linguistic patterns; a contextual phrase encoder captures nuanced semantic relationships; and a context-aware focal loss robustly mitigates the adverse effects of class imbalance. Crucially, unlike benchmark methods that often apply a static penalty to the majority class, our loss function dynamically modulates its focus based on the contextual difficulty and ambiguity of each individual sample. This allows for more precise handling of challenging borderline cases where standard loss functions typically falter. This study confirms that the integration of these specialized modules is critical for achieving robust performance, with the removal of any single component leading to a significant decline in the macro F1-score.

Beyond the technical achievement, SugarTextNet offers a novel analytical framework for social computing, demonstrating how targeted NLP techniques can uncover implicit societal issues and provide actionable tools for platform governance. Our findings underscore that bespoke architectural innovations, rather than simply scaling model size, are essential for addressing complex classification tasks in sensitive domains. The model's strong and balanced performance highlights its practical potential for deployment in online content moderation and risk detection systems where identifying rare but high-impact content is crucial.

\section{Discussion} 

\subsection{Translation Risks and Ethical Biases in Deployment}
When deploying such detection models, it is crucial to address ethical challenges and bias risks by adopting a value-neutral principle. The model should avoid moral judgments on consensual adult behaviors like "sugar dating" and prevent its misuse for social surveillance or reinforcing stigmas. Instead, its focus should be on protecting user privacy, preventing algorithmic discrimination, and mitigating associated risks such as financial disputes. Continuous auditing is necessary to correct biases stemming from cultural, generational, or privacy-related differences, thereby protecting the rights of minority groups.

\subsection{Additional Guidance on Deployment and Risk Mitigation}
To enhance the practical application effectiveness of the model and mitigate deployment risks, we recommend a multi-stage dynamic regulatory strategy. First, the model assesses content using multidimensional methods, including semantic keyword matching, homophone recognition, and user behavior history. Based on this assessment, a tiered management system is applied: "explicitly" violating content is immediately restricted, while "ambiguous" content is temporarily demoted (e.g., reduced visibility) and placed under continuous monitoring for re-evaluation using contextual signals like comment interactions. For content ultimately confirmed to be misjudged, restrictions are lifted, and traffic compensation may be offered to rectify the error. This entire process should be supported by a manual review mechanism and user appeal channels to ensure transparency and fairness.

\section{Limitations and Future Work}


While SugarTextNet shows strong performance, we identify three key limitations that guide our future work. First, the model's effectiveness, particularly for minority classes, is constrained by the dataset's scale. Future efforts will thus focus on expanding the dataset's size and diversity. Secondly, the model's scope is currently confined to the Chinese language and the Weibo platform. This restricts its generalizability, as Chinese social media (e.g., Xiaohongshu, Tieba) involves unique cultural specificities and metaphorical expressions not directly transferable to platforms predominantly using foreign languages (e.g., Twitter, Instagram). Therefore, a crucial next step is to develop multilingual detection mechanisms, likely by employing cross-lingual pre-trained models and leveraging transfer learning to enhance cross-platform adaptability. Finally, the model's text-only focus makes it vulnerable to evasive slang and multimodal content. We plan to address this by exploring multimodal fusion techniques that integrate features from images and videos to enhance robustness.It should also be noted that, due to computational and time constraints, this version does not include comprehensive statistical significance tests, detailed error analysis, or robustness evaluations—these will be prioritized in an extended journal version.

\printbibliography

\end{CJK*}
\end{document}